%% file: cross-modal-nature.tex
\documentclass[]{nature}

\usepackage{graphicx} 
\makeatletter
\let\saved@includegraphics\includegraphics
\AtBeginDocument{\let\includegraphics\saved@includegraphics}
\renewenvironment*{figure}{\@float{figure}}{\end@float}
\usepackage{mathptmx}
\makeatother
\usepackage{epstopdf}
\usepackage{caption}
\usepackage{subcaption}
\usepackage{floatrow}
\usepackage{multirow}
\usepackage{tabularx}
\usepackage{makecell}
\usepackage{courier}
\newcolumntype{C}[1]{>{\centering\let\newline\\\arraybackslash\hspace{0pt}}m{#1}}

\usepackage{macros}

\title{Cross-Modal Data Programming Enables Rapid Medical Machine Learning}

\author{Jared A. Dunnmon$^{1,*}$, Alexander J. Ratner$^{1,*}$,  Nishith Khandwala$^1$, Khaled Saab$^{2}$, Matthew Markert$^3$, Hersh Sagreiya$^4$, Roger Goldman$^4$, Christopher Lee-Messer$^3$, Matthew P. Lungren$^4$, Daniel L. Rubin$^{4,5,**}$, \& Christopher R\'{e}$^{1,**}$}

\begin{document}

\definecolor{keywords}{RGB}{255,0,90}
\definecolor{comments}{RGB}{0,0,113}
\definecolor{red}{RGB}{160,0,0}
\definecolor{green}{RGB}{0,150,0}

\newcommand{\mediumset}{Medium }
\newcommand{\largeset}{Large }
\newcommand{\demourl}{\texttt{www.github.com/HazyResearch/cross-modal-ws-demo}}

\lstset{language=Python, 
        basicstyle=\ttfamily\small, 
        keywordstyle=\color{keywords},
        commentstyle=\color{comments},
        stringstyle=\color{red},
        showstringspaces=false,
        identifierstyle=\color{green},
        procnamekeys={def,class}}

\maketitle

\begin{affiliations}
\item Department of Computer Science, Stanford University, Stanford, California, USA
\item Department of Electrical Engineering, Stanford University, Stanford, California, USA
\item Department of Neurology, Stanford University, Stanford, California, USA
\item Department of Radiology, Stanford University, Stanford, California, USA
\item Department of Biomedical Data Science, Stanford University, Stanford, California, USA
\end{affiliations}

\begin{abstract}
 \input{sections-nature/abstract}
\end{abstract}

 \input{sections-nature/intro}
 
 \input{sections-nature/applications}

 \input{sections-nature/results}

 \input{sections-nature/conclusion}

\begin{methods}
 \input{sections-supp/methods}
\end{methods}

\clearpage 
\include{sections-supp/statements}

\clearpage
\section*{References}
\bibliography{cross-modal-nature}

\clearpage
\include{sections-supp/supp-mat}

\end{document}

%% file: sections-nature/abstract.tex

Labeling \textit{training datasets} has become a key barrier to building medical machine learning models.
One strategy is to generate training labels programmatically, for example by applying natural language processing pipelines to text reports associated with imaging studies.
We propose \textit{cross-modal data programming}, which generalizes this intuitive strategy in a theoretically-grounded way that enables simpler, clinician-driven input, reduces required labeling time, and improves with additional unlabeled data.
In this approach, clinicians generate training labels for models defined over a target modality (e.g. images or time series) by writing rules over an auxiliary modality (e.g. text reports).
The resulting technical challenge consists of estimating the accuracies and correlations of these rules; we extend a recent unsupervised generative modeling technique to handle this cross-modal setting in a provably consistent way.
Across four applications in radiography, computed tomography, and electroencephalography, and using only several hours of clinician time, our approach matches or exceeds the efficacy of physician-months of hand-labeling with statistical significance, demonstrating a fundamentally faster and more flexible way of building machine learning models in medicine.

%% file: sections-nature/intro.tex

Modern machine learning approaches have achieved impressive empirical successes on diverse clinical tasks that include predicting cancer prognosis from digital pathology,\cite{bychkov2018deep,Yu2016} classifying skin lesions from dermatoscopy,\cite{Esteva2017} characterizing retinopathy from fundus photographs,\cite{Gulshan2016} detecting intracranial hemorrhage through computed tomography,\cite{Titano2018, Lee2018} and performing automated interpretation of chest radiographs.\cite{Dunnmon2018, Rajpurkar2018}
Remarkably, these applications typically build on standardized reference neural network architectures\cite{He2016} supported in professionally-maintained open source frameworks,\cite{abadi2016tensorflow,Paszke2017} suggesting that model design is no longer a major barrier to entry in medical machine learning.
However, each of these application successes was predicated on a not-so-hidden cost: massive hand-labeled training datasets, often produced through years of institutional investment and expert clinician labeling time, at a cost of hundreds of thousands of dollars per task or more.\cite{Gulshan2016,esteva2019guide}
In addition to being extremely costly, these training sets are inflexible: given a new classification schema, imaging system, patient population, or other change in the data distribution or modeling task, the training set generally needs to be relabeled from scratch.
These factors suggest that the biggest differentiator in medical machine learning today may be the collection of labeled training datasets,\cite{esteva2019guide} and point to a fundamental paradigm shift in how these applications are being built.

One manifestation of this shift in the broader machine learning community is the increasing use of \textit{weak supervision} approaches, where training data is labeled in noisier, higher-level, often programmatic ways, rather than manually by experts.
The statistical intuition behind these methods, which draw inspiration from a rich set of work on such topics as template-based systems for knowledge base creation,\cite{etzioni2004web} pattern-oriented bootstrapping for relation extraction,\cite{hearst1992automatic} and co-training techniques for leveraging unlabeled data,\cite{blum1998combining} is that a larger volume of noisier training data can sometimes be more effective than a smaller amount of hand-labeled data.
Classic examples of weak supervision approaches include using crowd workers,\cite{dawid1979maximum,Ghosh:2011:MMC:1993574.1993599,Dalvi:2013:ACB:2488388.2488414} heuristic rules,\cite{zhang2017position,zhang:cacm17,mallory2015large,halpern2014using,agarwal2016learning} distant supervision using external knowledge bases,\cite{mintz:acl09,craven:ismb99} coarse-grained image labels or multiple instance learning (MIL),\cite{mnih2012learning,wu2015deep,durand2017wildcat} and, more recently, end-to-end systems for managing and modeling programmatic weak supervision.\cite{RatnerAlexanderJandBachStephenHandEhrenbergHenryandFriesJasonandWuSenandRe2018Snorkel:Supervision}
In the medical imaging and monitoring domain, several recent approaches generate noisy training labels by applying natural language processing or text mining techniques to text reports accompanying the imaging or monitoring studies of interest.\cite{Titano2018,DBLP:journals/corr/WangPLLBS17,peng2018negbio,chilamkurthy2018deep}
We broadly characterize these methods as \textit{cross-modal weak supervision} approaches, in which the strategy is to programmatically extract labels from an \textit{auxiliary modality}---e.g. the unstructured text reports accompanying an imaging study---which are then used as training labels for a model defined over the \textit{target modality}, e.g. imaging studies.
These methods follow the intuition that programmatically extracting labels from the auxiliary modality can be far faster and easier than hand-labeling or deriving labels from the target modality directly.
However, these approaches still require substantial engineering effort for each new domain and task.

We propose \textit{cross-modal data programming}, a novel approach which aims to generalize, simplify, and theoretically ground the broad space of cross-modal weak supervision methods, while enabling us to study the efficacy of cross-modal weak supervision across multiple real applications.
In cross-modal data programming, rather than hand-labeling any training data, clinicians write \textit{labeling functions} (LFs), rules or heuristics which take in the auxiliary data point (e.g., a text report) and either output a label or abstain.
The goal is to then use these LF outputs to train a model defined over the target modality, such as a convolutional neural network (CNN) defined over 2-D radiograph inputs; in contrast to \textit{multi-modal} approaches, only this target modality is present at test time.
The fundamental assumption behind this approach is that writing LFs over the auxiliary modality is more accessible than writing LFs over the target modality; concretely, we assume that it is easier and faster to write LFs identifying mentions of a pathology (e.g. pneumothorax) in a report than to write LFs over the image that precisely identify that pathology.
These LFs can use pattern matching heuristics, complex clinical logic, other off-the-shelf classifiers, medical ontologies, or anything else that a clinician or scientist may consider to be relevant.
In this work, we study whether this remarkably simple and diverse programmatic input, generated with only hours of clinician time, can approach the efficacy of large hand-labeled training sets, annotated over months or years, for training machine learning models over complex target modalities including 2-D radiographs, 3-D volumetric imaging, and multi-channel time series.

This cross-modal data programming paradigm leads to a formal expression of the fundamental technical challenge that characterizes cross-modal weak supervision approaches: modeling the \textit{unknown} accuracies and correlations of different label sources (e.g. LFs), so as to properly combine and utilize their often overlapping and conflicting output labels.
We extend a recent unsupervised generative modeling technique,\cite{Ratner2016} and corresponding open-source Python package, Snorkel,\cite{RatnerAlexanderJandBachStephenHandEhrenbergHenryandFriesJasonandWuSenandRe2018Snorkel:Supervision} to handle this cross-modal setting in a provably consistent way.
Using this approach, we automatically estimate the accuracies and correlations of the clinician-authored LFs, re-weight and combine their outputs, and produce a final set of probabilistic training labels, which can then be used to supervise the model defined over the target modality.
Given mild statistical assumptions and LFs that are sufficiently pairwise independent according to a simple statistical test, we can guarantee that as more unlabeled data is used, the target modality model will increase in accuracy at the same asymptotic rate as if more hand-labeled data were being added.\cite{Ratner2019}
This core theoretical result implies that in settings where large amounts of unlabeled data are available, cross-modal data programming should be able to match or exceed the performance of models trained with large hand-labeled datasets.

We empirically test the hypothesis that cross-modal data programming can match---or even exceed---the efficacy of large hand-labeled datasets across four clinical applications: chest (CXR) and knee (EXR) radiograph triage, intracranial hemorrhage identification on head CT (HCT), and seizure onset detection on electroencephalography (EEG).
We find that with only a single day of clinician time spent writing LFs, our approach results in models that on average exceed the performance attained using physician-months of hand-labeled training data by 8.50 points area under the receiver operating characteristic curve (ROC-AUC), and on average achieve within 3.75 points ROC-AUC of the performance attained using physician-years of hand-labeled data.
Cross-modal data programming matches or exceeds the efficacy of physician-months of hand-labeling with statistical significance in each application.
Importantly, to provide supervision using this approach, clinicians wrote on average only 14 LFs per application, which comprised on average 6 lines of Python code each, and largely expressed simple pattern-matching or ontology-lookup heuristics; no extra effort was required to leverage additional unlabeled data.
Moreover, motivated by the increasing volume of digitized but unlabeled medical data, we evaluate the scaling of our approach with respect to additional unlabeled data, and find that performance consistently increases as more unlabeled data is collected in a way that is consistent with theoretical predictions.\cite{Ratner2016, Ratner2019}
In summary, cross-modal data programming can lower a substantial barrier to machine learning model development in medicine by serving as a fundamental new interface that reduces labeling time required from domain experts, thereby providing a stepping stone towards widespread adoption of machine learning models to provide positive, tangible clinical impact.

We first describe technical details of the cross-modal data programming approach.
We then provide an overview of the applications and datasets used. Finally, we discuss empirical results validating our hypotheses about speed and ease of clinician use, performance relative to large hand-labeled training datasets, and scaling with additional unlabeled data.

\paragraph{Cross-Modal Data Programming}
\begin{figure}[!tbp]
   \includegraphics[width=\textwidth]{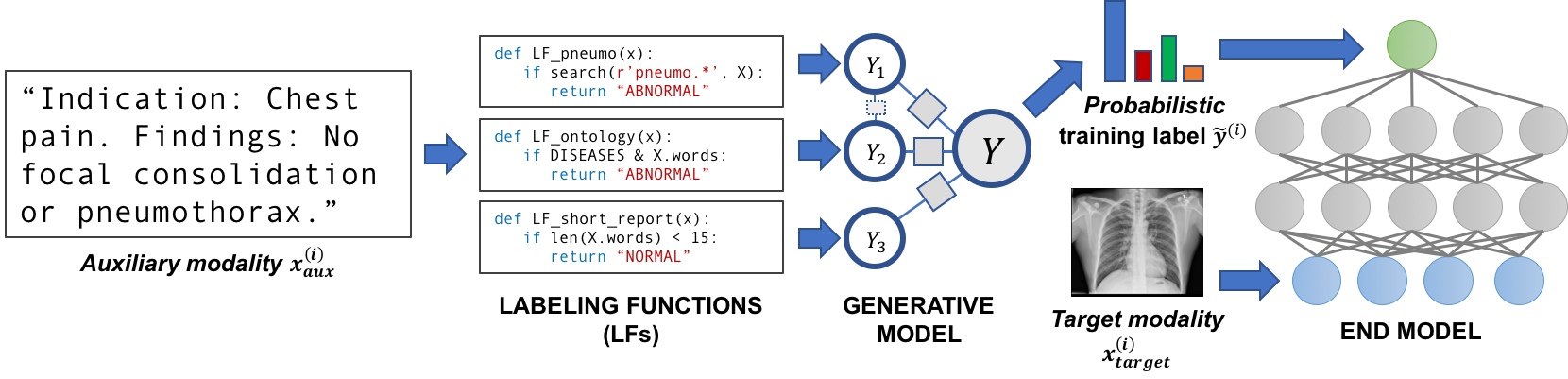}
   \caption{
 A \textit{cross-modal data programming} pipeline for rapidly training medical classifiers.
       A clinician first writes several \textit{labeling functions} (LFs), which are Python functions that express pattern-based rules or other heuristics, over the \textit{auxiliary} modality (e.g. text report).  
       Here, for instance, the function \textit{LF\_pneumo} would label a text report as ``abnormal" if the report contains a word with the prefix ``pneumo''; note that these LFs can be noisy, incorrect, and correlated with each other.
       During the offline model training phase, LFs are applied to unstructured clinician reports and combined to generate probabilistic (confidence-weighted) training labels for a classifier defined over the target modality (e.g. radiograph).
       At test time, the model receives only this target modality as input, and returns predictions.
   }   
   \label{fig:pipeline}
\end{figure}

In our proposed approach (Fig.~\ref{fig:pipeline}), clinicians provide two basic inputs: first, unlabeled cross-modal data points, which we represent as target-auxiliary modality pairs $(x^{(i)}_{\text{target}}, x^{(i)}_{\text{aux}}) \in \mathcal{X}_{\text{target}} \times \mathcal{X}_{\text{aux}}$ (e.g. an imaging study and the accompanying text report); and second, a set of  LFs, which are user-defined functions (e.g. pattern matching rules) that take in an auxiliary modality data point $x^{(i)}_{\text{aux}}$ as input, and either output a label, $\lambda^{(i)}_j \in \mathcal{Y}$, or abstain ($\lambda^{(i)}_j = 0$), where we consider the binary setting $\mathcal{Y} = \{-1,1\}$, corresponding to e.g. ``normal'' and ``abnormal.'' 
Importantly, our only theoretical requirement is that the majority of these LFs be more accurate than random chance.\cite{Ratner2016}
In our experiments, clinicians also hand-label a small (several hundred examples) \textit{development set} for use while writing LFs; however, this data is not used as training data to avoid potential bias.

Given $m$ LFs, we apply them to unlabeled auxiliary modality data points $\{x^{(i)}_{\textrm{aux}}\}_{i=1,\ldots,n}$ to generate a matrix of noisy labels, $\Lambda \in \mathbb{R}^{n \times m}$, where the non-zero elements in each row $\Lambda_i$ represent the possible training labels generated by the labeling functions.
In general these labels will overlap, conflict, and be arbitrarily correlated.
Our goal is to re-weight and combine them to generate a single probabilistic (i.e. confidence-weighted) label $\tilde{y}^{(i)}$.
To do this, we use an \textit{unsupervised generative modeling} procedure to learn both the correlation structure and accuracies of the LFs.\cite{Ratner2016,bach2017learning}
Concretely, we first estimate the structure of an exponential family generative model $p_\theta$, using an unsupervised technique,\cite{bach2017learning} and then learn the labeling function accuracies and correlation weights $\theta$ that best explain the observed labeling patterns by minimizing the negative log marginal likelihood:
\begin{align}
   \hat{\theta}
   &=
   \textrm{argmin}_{\theta} \left( -\log\sum_{Y} p_\theta(Y, \Lambda) \right)
   \label{eq:gm}
\end{align}
where $Y$ is the unobserved true label vector over which we marginalize.

The result of this first modeling stage is a set of \textit{probabilistic} or confidence-weighted training labels $\tilde{y}^{(i)}$.
We then use these to train a \textit{weakly-supervised discriminative model} $f_w$ (e.g. a neural network) over the target modality, using a \textit{noise-aware} variant of the loss function $l$,\cite{Ratner2016} which takes into account the uncertainty of the probabilistic training labels:
\begin{align}
   \hat{w}
   &=
   \textrm{argmin}_w \sum_{i=1}^n \mathbb{E}_{\tilde{y}^{(i)} \sim p_\theta(\cdot|\Lambda_i)}\left[ l(f_w(x_{\textrm{target}}^{(i)}), \tilde{y}^{(i)}) \right]
   \label{eq:dm}
\end{align}
The resulting model---represented by the estimated parameters $\hat{w}$---can then be applied to the target modality alone; for example, to classify chest radiographs by triage priority before human interpretation.
Importantly, we rely on recent statistical learning theory\cite{Ratner2019} which guarantees that, under certain basic statistical assumptions, the estimation error of the learned LF accuracies and correlation parameters (Equation~\ref{eq:gm}) and the test error of the discriminative model (Equation~\ref{eq:dm}) will both be bounded by $O(n^{-\frac12})$.
That is, as the size $n$ of the \textit{unlabeled} dataset increases, we should observe improved performance at the same asymptotic rate as in traditional supervised approaches when adding more hand-labeled samples.
This result provides a theoretical framework for understanding the broad spectrum of cross-modal weak supervision methods that can be formulated in our approach, and how they can leverage both clinician domain expertise and available unlabeled data.

%% file: sections-nature/applications.tex

\section*{Datasets and Experimental Procedure}

\begin{figure*}
    \centering
    \begin{subfigure}[b]{0.475\textwidth}
        \centering
        \includegraphics[width=0.93\textwidth]{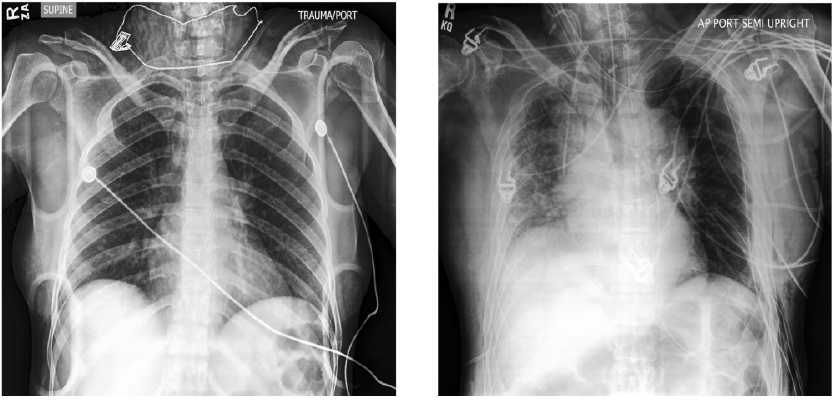}
        \caption[CXR]%
        {{\small \textbf{CXR:} Example normal (left) and abnormal (right) chest radiographs.  Note view is the same.
        }}    
        \label{fig:results-data-cxr}
    \end{subfigure}
    \hfill
    \begin{subfigure}[b]{0.475\textwidth}  
        \centering 
        \includegraphics[width=0.93\textwidth]{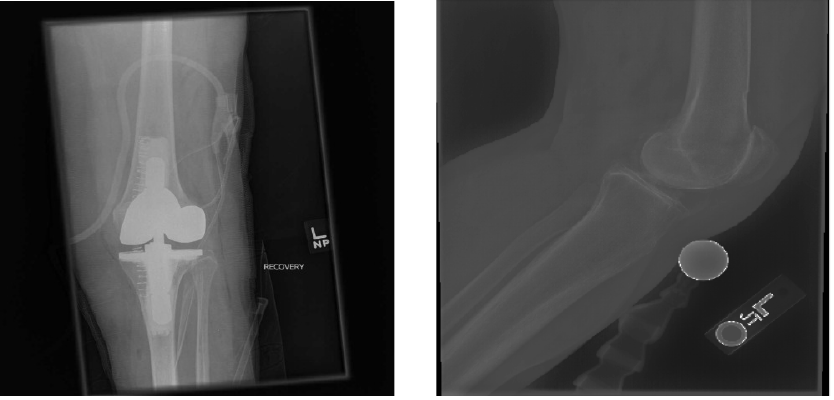}
        \caption[MSK]%
        {{\small \textbf{EXR:} Example normal (left) and abnormal (right) knee radiographs.  Note view is different.
        }}    
        \label{fig:results-data-msk}
    \end{subfigure}
    \vskip\baselineskip
    \begin{subfigure}[b]{0.475\textwidth}   
        \centering 
        \includegraphics[width=\textwidth, ]{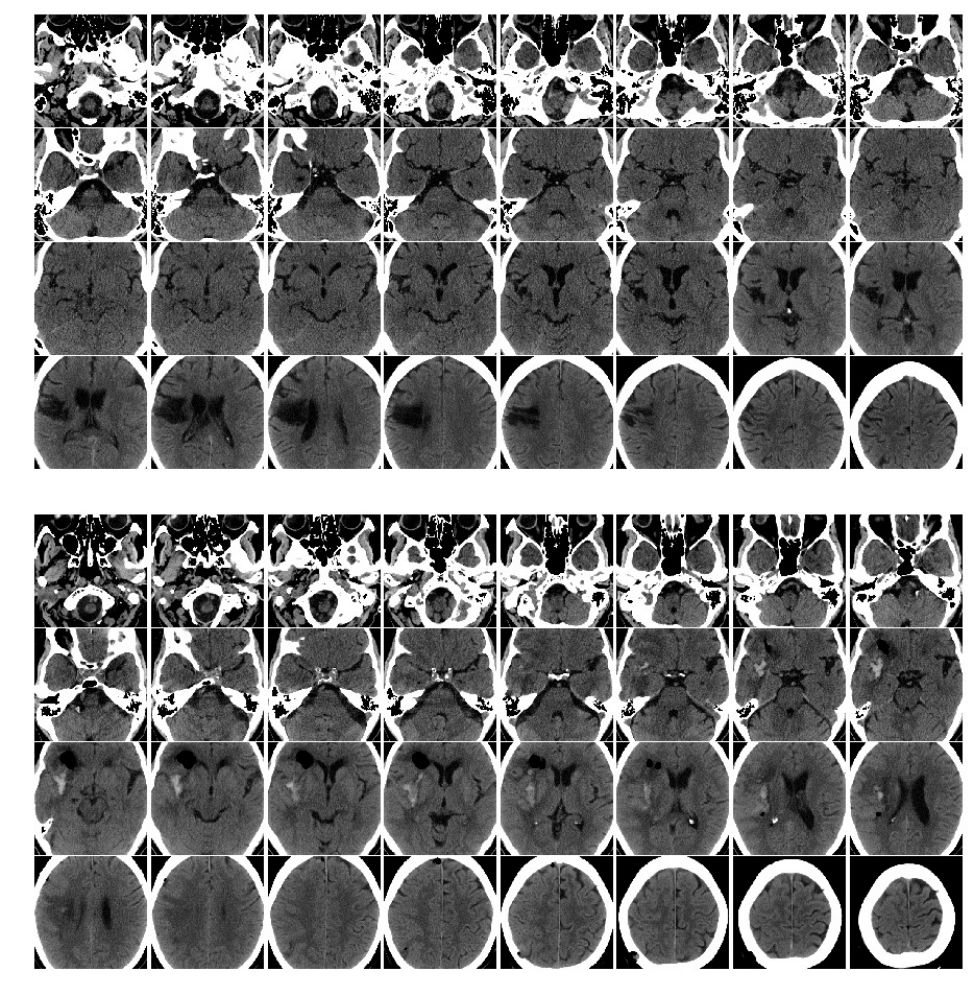}
        \caption[HCT]%
        {{\small \textbf{HCT:} Example HCT signals denoting no hemorrhage (top) and hemorrhage (bottom).}}    
        \label{fig:results-data-hct}
    \end{subfigure}
    \quad
    \begin{subfigure}[b]{0.475\textwidth}   
        \centering 
        \includegraphics[width=0.93\textwidth]{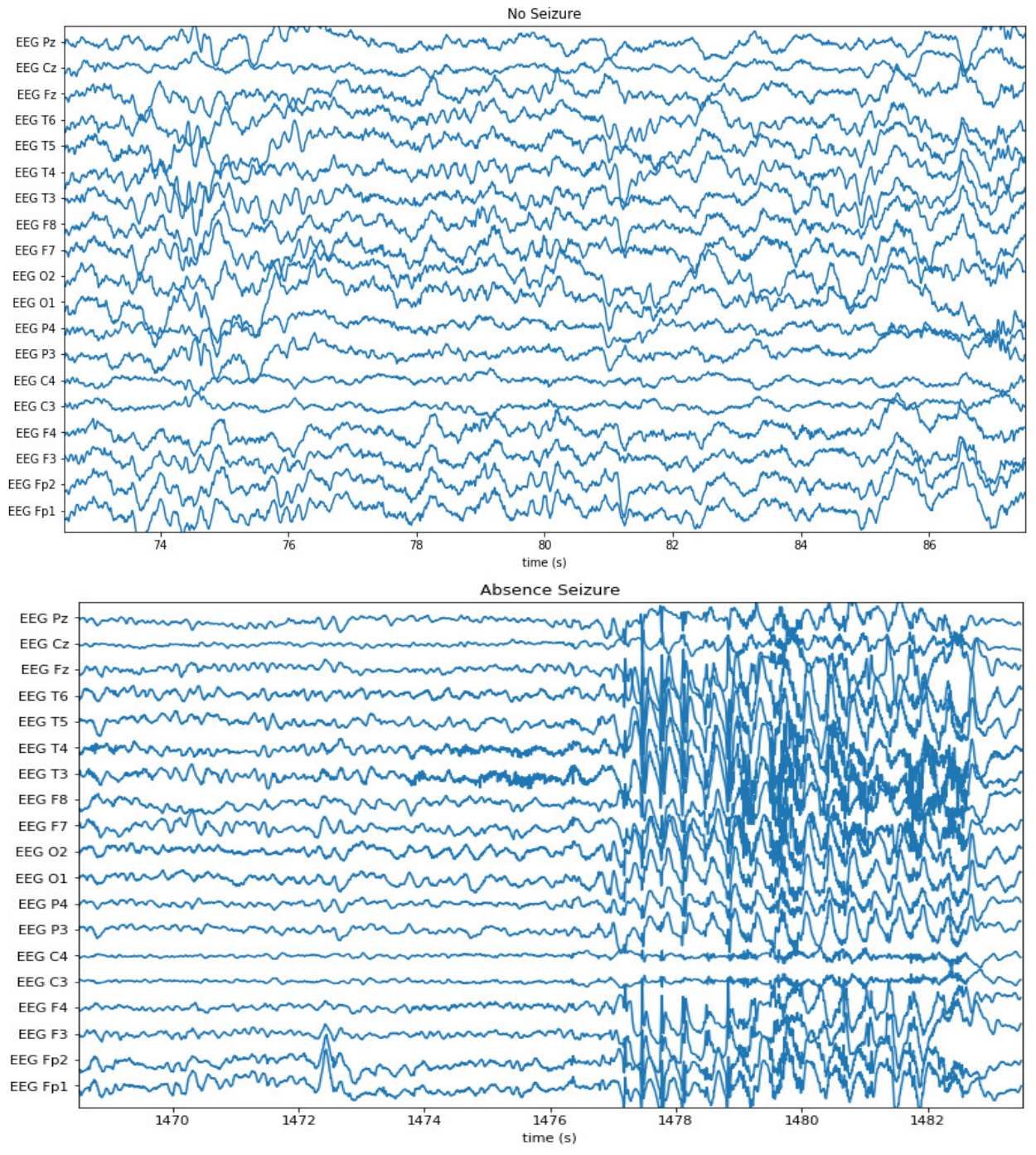}
        \caption[EEG]%
        {{\small \textbf{EEG:} Example EEG signals denoting no seizure (top) and seizure onset (bottom).}}    
        \label{fig:results-data-eeg}
    \end{subfigure}
    \caption[data]%
    { Example target modality data for the four applications surveyed, which demonstrates the breadth of applicability of the proposed cross-modal weak supervision approach; auxiliary modality data (text reports) not pictured.  Panel (a) shows single 2-D chest radiographs (CXR), panel (b) shows examples of knee extremity radiographs (EXR) drawn from 2-D radiograph series, panel (c) shows 32 slices from 3-D head CT scans (HCT) with and without hemorrhage, and panel (d) shows 19-channel electroencephalography (EEG) signals with and without evidence of seizure onset.  Note that while these applications are fundamentally different in both dimensionality and modeling requirements, as described in Table~\ref{table:app-table}, deep machine learning models supporting each can be rapidly trained using the cross-modal data programming technique we present here.}
    \label{fig:results-data}
\end{figure*}

We assess the performance of the proposed cross-modal data programming approach, compared to using hand-labeled data, in four real-world medical applications  (Fig. \ref{fig:results-data}) spanning chest radiograph triage (2-D image classification), extremity radiograph series triage (2-D image series classification), intracranial hemorrhage detection on CT (3-D volumetric image classification), and seizure onset detection on EEG (19-channel time series classification), using commodity deep neural network architectures for each (Table \ref{table:app-table}).  
In order to provide a rigorous comparison between cross-modal data programming and hand-labeled data in our experiments, we curate a large hand-labeled dataset for each application comprising raw data, associated reports, and clinician-provided labels; each of these datasets represents physician-months or physician-years of hand-labeling.  
Using these data resources, we can assess how closely the performance of models trained using cross-modal data programming can come to matching their fully supervised equivalents in the context of real clinical data.  
For each application, we use the cross-modal data programming approach as described above, and as implemented as an extension to the Snorkel\cite{RatnerAlexanderJandBachStephenHandEhrenbergHenryandFriesJasonandWuSenandRe2018Snorkel:Supervision} software package: a clinician first writes labeling functions over the text reports; we then use Snorkel to generate a final set of probabilistic training labels; and we finally train a discriminative model over the target data modality.
We describe this process in detail below for HCT, deferring details of other applications to the Methods section.  

We first curate and preprocess the dataset used for each application. 
For HCT, we create a dataset describing the binary task of intracranial hemorrhage detection by collecting 5,582 non-contrast HCT studies from our institution's Picture Archiving and Communications System (PACS) and procuring their associated text reports. 
We restrict our dataset to studies containing between 29 and 45 axial slices reconstructed at 5 mm axial resolution and retain the center 32 slices of each reconstruction, padding with images containing values of 0 Hounsfield Units where necessary. 
Clinicians then hand-label a small \textit{development set}, which was used not as training data, but rather as an aid for tuning both clinician-provided LFs and model hyperparameters; for HCT, this required clinicians to label 170 HCT studies as positive or negative for hemorrhage.

As a second step, clinicians write LFs that for each report either provide a label or else abstain. 
For HCT, a single radiology fellow composed seven Python LFs over the text report based on their own experience reading and writing radiology reports, with several hours of support from a computer science graduate student.
Third, we train a generative model to simultaneously learn the accuracies of all LFs. 
Concretely, we compute the $\Lambda$ matrix by executing LF code to calculate output values for $m=7$ LFs on $n=4,000$ training examples, and then execute a single Python command in Snorkel\cite{RatnerAlexanderJandBachStephenHandEhrenbergHenryandFriesJasonandWuSenandRe2018Snorkel:Supervision} to estimate generative model parameters $\hat{\theta}$ by solving Eq.~(\ref{eq:gm}).
Fourth, we assign a composite probabilistic label that represents an appropriately weighted combination of the LF outputs; practically, this translates into executing the trained generative model over each report.
Clinicians then compare the output of the generative model to their ground truth development set labels, and repeat steps 2 - 4 until diminishing returns are observed with respect to generative model performance as evaluated against the development set.
This entire procedure generally requires fewer than eight cumulative hours of clinician time per task.
 
Once weak labels have been provided for a given task, a discriminative machine learning model (e.g. neural network classifier) can be trained over the raw data modality to evaluate the quantity of interest. 
For HCT, we define a standard attention mechanism over an 18-layer Residual network\cite{He2016} encoder in PyTorch\cite{Paszke2017} that operates on every axial slice of the reconstructed tomographic image; this attention mechanism is a small neural network that dynamically learns how heavily a given slice should be weighted in the representation used by the final classification layer.\cite{Ilse2018}  
We then estimate optimal parameters $\hat{w}$ by approximating the solution to Eq.~(\ref{eq:dm}) using standard PyTorch backpropagation algorithms and a binary cross entropy loss function between the network prediction and the weak probabilistic label.
Implementation details for each application are provided in Methods, while LF code for each and a functional demonstration of the entire technique on a small, public dataset is available online.\footnote{\demourl}

\begin{table}
\centering
\begin{tabularx}{17.45cm}[!tbh!]{|C{2.75cm}|C{2cm}|C{3.5cm}|C{4cm}|C{3cm}|}
\hline
  & CXR & EXR & HCT & EEG   \\
\hline
\hline
Data Type & Single 2-D \newline Radiograph & Multiple 2-D \newline Radiograph Views & 3-D CT \newline Reconstruction & 19-Channel EEG \newline Time Series \\
\hline
Task &  Normal \newline Abnormal & Normal \newline Abnormal & Hemorrhage \newline No Hemorrhage & Seizure Onset \newline No Seizure Onset \\
\hline
Anatomy & Chest & Knee & Head & Head \\
\hline
Train Set Size \newline 
(Large/Medium)  &  50,000 \newline 5,000 & 30,000 \newline 3,000 & 4,000 \newline  400 & 30,000 \newline 3,000 \\
\hline
Train Set Size \newline (Literature) & 20,000\cite{Dunnmon2018}  &  40,561\cite{Rajpurkar2017} & 904\cite{Lee2018} &  23,218\cite{Obeid2016} \\
\hline
Network \newline Architecture & 2-D \newline ResNet-18\cite{He2016} & Patient-Averaged \newline 2-D ResNet-50\cite{He2016} & 3-D MIL + ResNet-18 \newline + Attention\cite{Ilse2018} & 1-D Inception \newline DenseNet\cite{Roy2018}\\
\hline
\hline
\end{tabularx}
\caption{Description of data type, classification task, train set sizes, and neural network architectures for each application studied.  We evaluate cross-modal data programming on four different data types: 2-D single chest radiographs (CXR), 2-D extremity radiograph series (EXR), 3-D reconstructions of computed tomography of the head (HCT), and 19-channel electroencephalography (EEG) time series.  Tasks are each binary for simplicity, though our approach also handles higher cardinality classification problems.  We present two different dataset sizes used in this work: the full labeled dataset (Large) of a size that might be available for an institutional study (i.e. physician-years of hand-labeling), and a 10\% subsample of the entire dataset (Medium) of a size that might be reasonably achievable by a single research group (i.e. physician-months of hand-labeling).  
For context, we present the size of comparable datasets recently used to train high-performance models in the literature for each domain in the ``Train Set Size (Literature)" field.  
Finally, we list the different commodity model architectures used, demonstrating the diversity of compatible models.  
While each image model uses a Residual network as an encoder,\cite{He2016} architectures vary from a simple single-image network (CXR) to a mean across multiple image views (EXR) to a dynamically weighted attention mechanism that combines image encodings for each axial slice of a volumetric image (HCT).  For time series, an architecture combining the best attributes of the Residual and Densely Connected\cite{Huang2016} networks for 1-D applications is used, where each channel is encoded separately and a fully-connected layer is use to combine features extracted from each (see Methods).  }
\label{table:app-table}
\end{table}

\label{sec:eeg}

%% file: sections-nature/results.tex

\section*{Results and Discussion}
\label{sec:results}
We empirically assess our hypothesis that cross-modal data programming can enable fast construction of machine learning applications across diverse clinical settings along three principle axes: the performance relative to large hand-labeled datasets, the scaling of performance with additional unlabeled data, and the time and complexity of required clinician input.

\subsection{Weakly Supervised Performance Matches or Exceeds Fully Supervised Performance}

\begin{figure}[t!]
    \includegraphics[width=6 in]{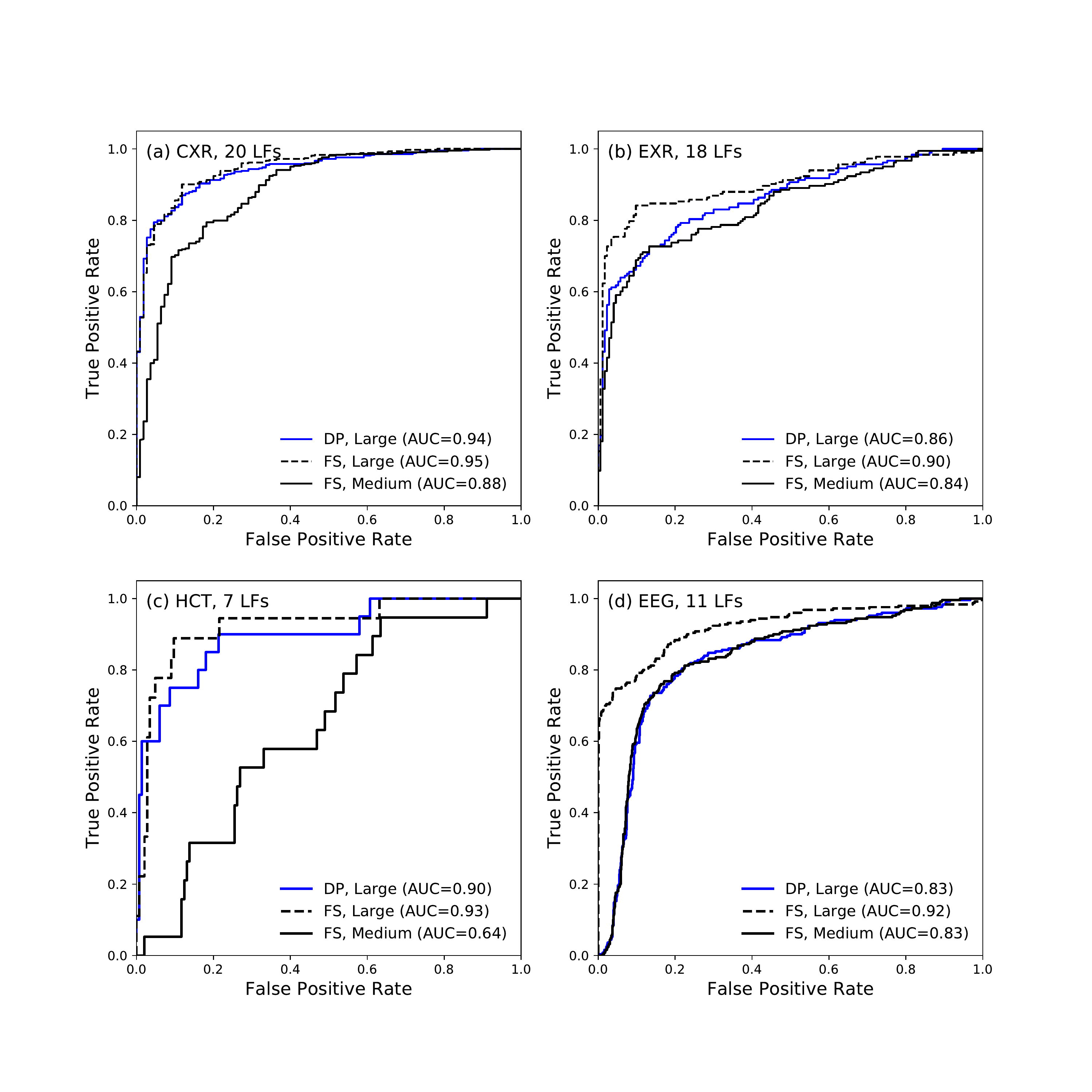}
    \caption{ROC curves for models trained using full hand-labeled supervision (FS) and cross-modal data programming (DP) for (a) chest radiographs (CXR), (b) extremity radiographs (EXR), (c) head CT (HCT), and (d) electroencephalography (EEG).  Each curve shown is that attaining the median ROC-AUC score on the test set with respect to model training runs using five different random seeds (see Methods).  In each case, models trained using cross-modal data programming demonstrate performance levels that meet or exceed those of models trained using \mediumset fully supervised datasets (i.e. physician-months of labeling time), and approach those of models trained using \largeset fully supervised datasets (i.e. physician-years of labeling time).  See Table \ref{table:app-table} for additional details on dataset sizes.  Details of statistical analysis can be found in Methods, with results of statistical tests reported in Extended Data.}
    \label{fig:fig-roc}
\end{figure}

We first assess whether cross-modal weak supervision can replace large hand-labeled training sets with flexible and high-level programmatic supervision specified by subject matter experts.  
In Fig.~\ref{fig:fig-roc}, we compare cross-modal data programming (DP) to full hand-labeled supervision (FS) with \largeset and \mediumset hand-labeled datasets (see Table \ref{table:app-table} for dataset size definitions), finding that across applications, median models  (i.e. models achieving median ROC-AUC across five random seeds) trained using cross-modal data programming are able to perform within an average of $3.75$ ROC-AUC points of those trained with \largeset hand-labeled datasets while outperforming those trained using \mediumset hand-labeled datasets by an average of $8.50$ ROC-AUC points.  
Median weakly supervised models for CXR and HCT are not statistically different than median models trained using \largeset hand-labeled datasets (DeLong p$>$0.3),\cite{DeLong1988} while median weakly supervised EXR and EEG models are not statistically different than median models trained using \mediumset hand-labeled datasets (DeLong p$>$0.25). 
All relevant p-values are provided in Extended Data.
We view the impressive empirical performance of the cross-modal weakly-supervised models as compared to the fully-supervised ones as a notably positive outcome of our study, and in turn, view the annotation and curation effort required to perform this comparison between weakly and fully supervised models as a major contribution of this work.
Because relative rather than absolute performance assessments are the goal of the study, we used standard model architectures and a coarse hyperparameter search for each problem (see Methods); nonetheless, our fully supervised results compare favorably to recently published work on each problem, reflecting encouragingly on the potential clinical utility of analogous weakly supervised models.\cite{Dunnmon2018, Rajpurkar2017,Titano2018,Lee2018,Acharya2018}

\subsection{Models Trained Using Cross-modal Data Programming Improve with More Unlabeled Data}

\begin{figure}[!tbp!]
    \includegraphics[width=6 in]{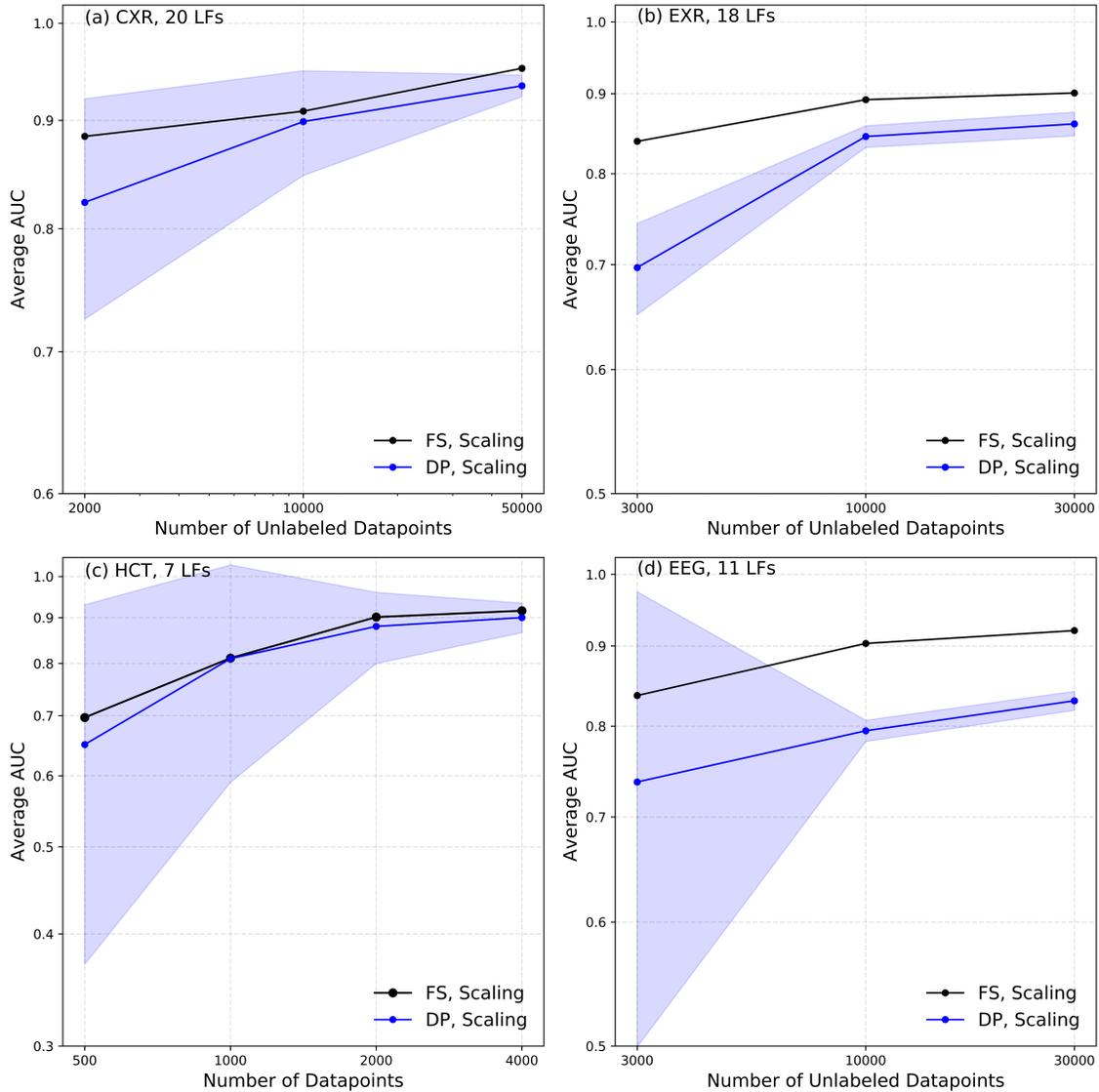}
    \caption{Mean neural network ROC-AUC score vs.~dataset size using full supervision (FS) and cross-modal data programming (DP) for (a) chest radiographs (CXR), (b) extremity radiographs (EXR), (c) head CT (HCT), and (d) electroencephalography (EEG).  In each case, models trained using cross-modal data programming show similar scaling properties to those trained using full hand-labeled supervision.  Error bars represent 95\% confidence intervals from five training runs with different random seeds.}
    \label{fig:fig-scaling}
\end{figure}

A major trend in the modern clinical world is that large amounts of digitized medical data are increasingly accessible, but not labeled in the correct format for supervising machine learning models; we therefore evaluate whether cross-modal data programming can harness this growing data resource without requiring additional clinician effort.
Specifically, we assess the performance of the target modality model when additional unlabeled data is weakly labeled using cross-modal data programming, but without any modification of the fixed set of clinician-authored LFs.  
In Fig.~\ref{fig:fig-scaling}, we observe consistent performance scaling across all applications studied.

In particular, the chest radiograph results of Fig.~\ref{fig:fig-scaling}a demonstrate that the weakly supervised model achieves similar asymptotic ROC-AUC scaling to the fully supervised model, as suggested by recent theoretical analysis.\cite{Ratner2016,Ratner2019}
Fig.~\ref{fig:fig-scaling}b describing extremity radiograph triage shows analogous scaling trends, but the initial improvement in performance of the weakly supervised model upon adding more data points appears far more rapid.
We speculate that this is due to the combination of weak supervision with a relatively simple MIL modeling approach (i.e. mean across radiographs), and propose that this phenomenon is worthy of further investigation.
These scaling trends also hold for more complex modalities in HCT and EEG (Fig.~\ref{fig:fig-scaling}c,d).
Indeed, we not only observe rapid mean performance improvement in the weakly supervised CT and EEG models with additional examples, but also demonstrate a substantial reduction in model variance across training runs with different random seeds.
Thus, our scaling results suggest that machine learning models supervised using cross-modal data programming---and requiring no additional clinician effort to harness additional available data---can asymptotically approach the fully supervised result not only in theory, but also in practice on a variety of real-world clinical datasets.

\subsection{Cross-Modal Weak Supervision Requires Simple and Minimal Clinician Input}

\begin{figure}[!tbp]
    \centering
    \includegraphics[width=4 in]{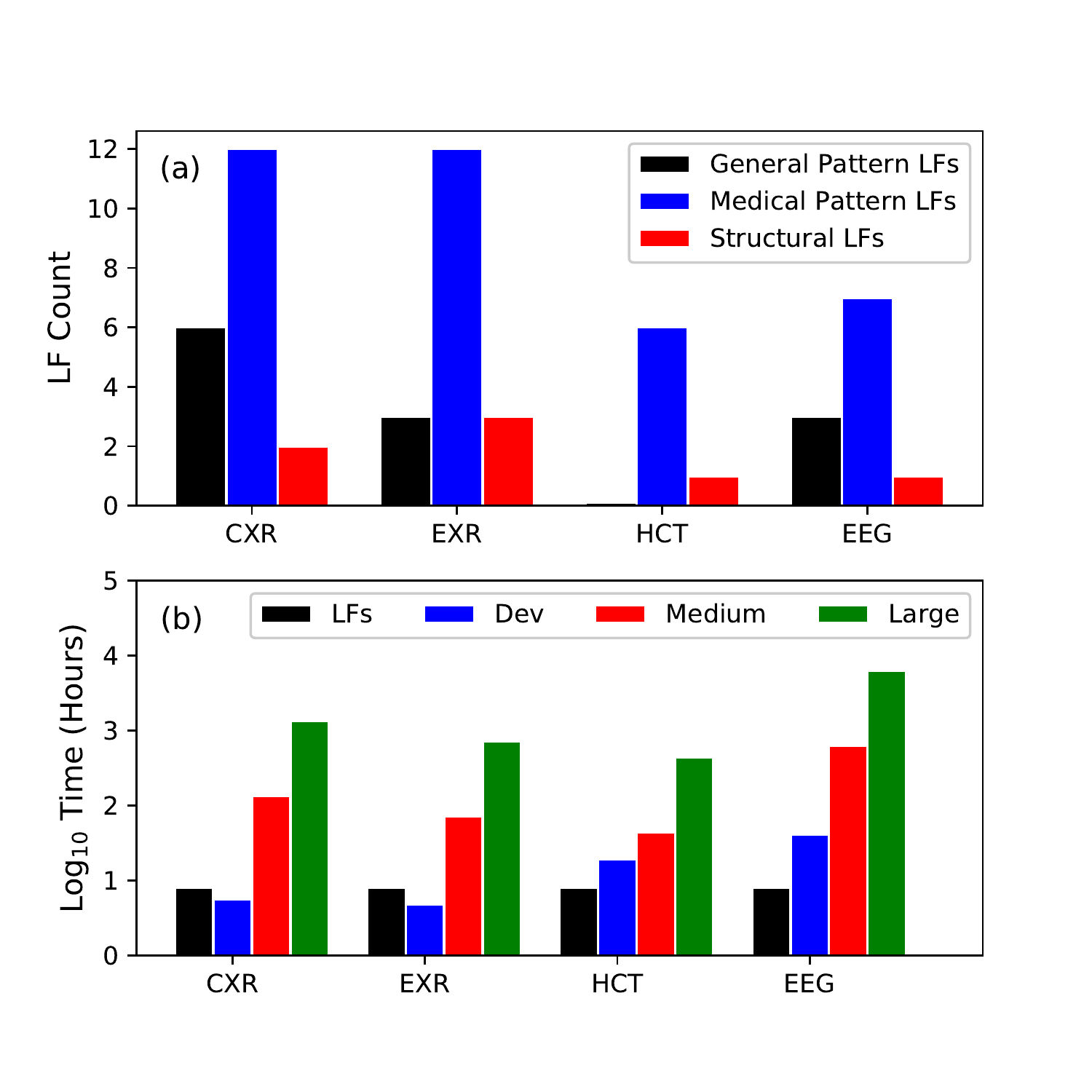}
    \caption{Labeling function (LF) types for each application and labeling time (log scale) for fully (hand-labeled) and weakly (cross-modal data programming) labeled datasets containing chest radiographs (CXR), extremity radiographs (EXR), head CT (HCT), and electroencephalography (EEG).  Labeling times are presented for the small development set (Dev) of several hundred examples, the full labeled dataset (Large) of a size that might be available for an institutional study (i.e. physician-years of labeling time), and a 10\% subsample of the entire dataset (Medium) of a size that might be reasonably achievable by a single research group (i.e. physician-months of labeling time). See Table \ref{table:app-table} for additional details on dataset sizes.  Because the hand-labeled development set would be required for both weakly and fully supervised model training, it is simply useful to observe that time required to label the development set is usually of a similar order of magnitude to that required to write the LFs.  Labeling times for fully supervised datasets were estimated using the overall dataset size combined with median read times of 1 minute 38 seconds per CXR, 1 minute 26 seconds per EXR, 6 minutes 38 seconds per HCT, and 12 minutes 30 seconds per EEG that are drawn directly from reported values in the literature.\cite{Cowan2013,Brogger2018} Note that the hand-labeling timing estimates presented here are conservative in the sense that they account only for a single annotator, whereas medical machine learning studies often rely on multiple-annotator data.\cite{Gulshan2016, Esteva2017, Lee2018}} 
    \label{fig:fig-timing}
\end{figure}

Finally, we analyze the types and amount of input that were required of clinicians in our cross-modal data programming approach relative to hand-labeling data.
On average, 14 labeling functions comprising on average 6 lines of code each were developed per application, using a mix of \textit{general text patterns} (e.g. identifying non-medical words indicating normalcy, equivocation, etc.), \textit{medical text patterns} (e.g. identifying specific terms or overlaps with medical ontologies), and \textit{structural heuristics} (e.g. shorter reports tend to describe normal cases), as described in Figure~\ref{fig:fig-timing}(a). 
Importantly, between the CXR and EXR applications 6 LFs were directly reused, demonstrating the ability of our approach to amortize clinician efforts across different modeling tasks.
While composing LFs took less than a single person-day per application, hand-labeling training sets required physician-months to physician-years (Figure~\ref{fig:fig-timing}(b)).
Thus, we see not only that cross-modal data programming achieves encouraging performance results, but also that it does so while requiring orders of magnitude less time spent on the labeling process.

\subsection{Limitations}  Several limitations should be considered when interpreting the results of this study.  First, because we use data sourced from a single hospital, our results do not address cross-institutional model validation.  Second, though the statistical estimation techniques underlying cross-modal data programming do handle the multi-class setting, this work focuses on well-defined binary tasks.  Third, this technique is limited to cases wherein an auxiliary modality exists, is readily available, and is amenable to rapid LF specification. Finally, our study assumes the existence of reference neural network architectures for a given diagnostic modality; while this is often valid in computer vision, additional resources could be required to create high-performance models for other diagnostic modalities.  Addressing each of these points would be a natural and valuable direction for future research.

%% file: sections-nature/conclusion.tex

\section*{Conclusion}
In this work, we have described a new paradigm of weakly supervising medical machine learning problems via \textit{cross-modal data programming}.
Further, we have demonstrated that it can be a powerful and broadly-applicable technique for training modern machine learning models in the context of complex clinical data modalities as varied as radiography, computed tomography, and electroencephalography, when auxiliary modalities amenable to clinician heuristic supervision, such as text reports, are available.
Our results show that this approach can enable clinicians
to rapidly generate large training sets in a matter of days which match or exceed the performance of large, hand-labeled training sets that took months or years of phyisican-time to annotate.
This approach has the potential to standardize many current application-specific approaches to assembling labeled training sets, is flexible across diverse use cases, requires relatively little machine learning expertise, and is compatible with nearly any downstream machine learning model.
Furthermore, we demonstrated that our cross-modal data programming approach can leverage the rising tide of available unlabeled clinical data along with recent advances in unsupervised generative modeling techniques to combine clinician-provided inputs in a way that rapidly provides programmatic labels of sufficient quality to support training of highly performant machine learning models.\cite{Ratner2016,RatnerAlexanderJandBachStephenHandEhrenbergHenryandFriesJasonandWuSenandRe2018Snorkel:Supervision}
More broadly, by shifting to programmatically-generated training data that leverages domain expertise, this approach enables rapid modification, maintenance, and transfer of training sets to new modeling targets, annotation schemas, or datasets.
It is our hope that this work represents a stepping stone towards higher-level, more practical paradigms of generating training data for modern machine learning models in the clinical setting, ultimately leading to the widespread adoption of scalable, human-in-the-loop machine learning pipelines with the potential for positive, tangible clinical impact.

%% file: sections-supp/methods.tex

Each dataset analyzed here was retrospectively collected from our institution.  Institutional Review Board (IRB) approval was obtained in each case.  Details of dataset composition, preprocessing routines, training procedures, and model architectures are provided for each application.  In each dataset, different patients were assigned to train, validation, and test sets.  For each commodity model architecture used, hyperparameters were determined via coarse grid search.   We note that further optimization could have been performed with respect to both architecture and hyperparameter search; while such fine-tuning tasks can and should play an important role in translating academic results into potential deployment, they are not a focus of this study. 

\subsection{Automated Triage of Frontal Chest Radiographs}
As the demand for imaging services increases, automated triage for common diagnostics such as chest radiographs is expected to become an increasingly important part of radiological workflows.\cite{Titano2018, Dunnmon2018}
Our frontal chest radiography dataset comprises a 50,000-sample subset of the automated triage dataset described in Dunnmon et al.\cite{Dunnmon2018} containing paired images and text reports, where each example describes a unique patient.  Each image is also associated with a prospective normal or abnormal label provided by a single radiologist at the time of interpretation, and the dataset balance is 80\% abnormal.  
The fully supervised (hand-labeled) results are trained using these prospective labels.  
For both fully and weakly supervised models, we use a development set size of 200 prospectively labeled images for cross-validation during the training process and evaluate on the same 533-image held-out test dataset as Dunnmon et al., which is labeled by blinded consensus of two radiologists with 5 and 20 years of training, for the sake of consistency with the published literature.\cite{Dunnmon2018}   Examples of chest radiographs can be found in Fig.~\ref{fig:results-data-cxr}.

For this task, a radiology fellow wrote 20 LFs over the text in under 8 hours of cumulative clinician time, using a labeled development set of 200 reports.
These LFs were implemented in Python with several hours of assistance from a computer science graduate student.  
As is usual in the data programming paradigm, these functions represent a combination of pattern-matching, comparison to known knowledge-bases, and domain-specific cues. 
 In the context of chest radiography, the impression and findings sections of the radiology report tend to be the most informative; thus, the majority of LFs considered text within these sections.  
  Once LFs have been generated, tools within the Snorkel software package were used to generate a set of probabilistic training labels.\cite{RatnerAlexanderJandBachStephenHandEhrenbergHenryandFriesJasonandWuSenandRe2018Snorkel:Supervision}
 Note that this particular set of LFs is not meant to be prescriptive or exhaustive; rather, it represents the output of a real-world effort to create useful rules for programmatic labeling of data with domain experts in a reasonable period of time.  
 While we would not expect models that perform in exactly the same manner if different radiologists were to compose LFs for this task, previous work has demonstrated that the unsupervised statistical estimation techniques underlying cross-modal data programming are able to effectively combine noisy labeling signal in a way that improves model performance across a large number of users.\cite{RatnerAlexanderJandBachStephenHandEhrenbergHenryandFriesJasonandWuSenandRe2018Snorkel:Supervision} 

The image model used for this task was an 18-layer Residual Network (ResNet-18) with a sigmoid nonlinearity on top of a single-neuron final layer, which yields near-state-of-the-art results on this dataset at a modest computational cost.\cite{He2016,Dunnmon2018}
The model was implemented using the PyTorch software framework, and was initialized using weights pre-trained on the ImageNet database.\cite{Paszke2017,JiaDeng2009} 
 Following Dunnmon et al.,\cite{Dunnmon2018} models were trained on a single Tesla P100 GPU using the Adam optimizer with default parameters and early stopping, an initial learning rate of 0.001, batch size of 72, learning rate decay rate of $\sqrt{0.1}$ on plateau in the validation loss, weight decay (i.e. $L_2$ regularization) value of $0.005$, and the binary cross-entropy loss function. 
 Images were preprocessed using histogram equalization, downsampled to 224 x 224 resolution, z-score normalized using global mean and standard deviation values computed across the dataset, and replicated over three channels (for compatibility with a model originally for RGB images) before injection into the training loop.  
 Each model over 50,000 images took approximately six hours to train.  

Because each image is associated with a unique report, evaluation is performed on a simple image-by-image basis using the area under the receiver operating characteristic curve (ROC-AUC) as the evaluation metric.  
Given this setup, we do not require cutoff tuning because the ROC-AUC metric integrates performance across all possible cutoffs.  
Values are reported in graphical form using ROC-AUC as a performance metric, with 95\% confidence intervals on the ROC-AUC values derived from five separate runs of each model for each dataset size using different random seeds for train-development splitting and other randomized parameters.  The held-out test set was the same in all cases.

\subsection{Automated Triage of Extremity Radiograph Series}
Musculoskeletal disorders are generally diagnosed from imaging, and their substantial prevalence makes them a natural target for automated triage systems; in this work, we specifically focus on detecting musculoskeletal disorders of the knee from multiple-view radiograph series.
We originally obtained a dataset of 3,564 patient exams prospectively labeled as normal or abnormal in the same manner as described for chest radiography, containing a total of 37,633 individual images and 3,564 reports.
A single exam can have a variable number of radiographs across different views, as collection protocols are not standard for this type of exam.  
The dataset was pulled from the PACS system in such a way that it contains a 50-50 distribution of normal and abnormal cases (as indicated by the prospective label). 
We split the dataset such that images from a randomly sampled 3,008-patient cohort (approximately 30,000 radiographs) are available for training both fully supervised and weakly supervised models, with images from 200 patients available for cross-validation.  
We then obtained retrospective labels for each image in a held-out test set of 356 patient exams (3,718 images), where a single radiologist with 9 years of training provides a normal or abnormal label with no time constraints.  
We used this set of 356 exams for evaluation on a patient-by-patient basis.  
Examples of knee radiographs can be found in Fig.~\ref{fig:results-data-msk}.

For this task, a single radiology fellow wrote the 18 LFs within eight hours of cumulative clinician time, which were implemented in Python with several hours of assistance from a computer science graduate student using a development set of 200 labeled reports.
These LFs use a variety of pattern-matching and semantic cues to identify reports as either normal or abnormal.  
While there is some overlap between these LFs and those for the chest radiographs (e.g. the length of the report), LFs nonetheless had to be tuned for this specific application.
Probabilistic labels were then generated from these LFs using tools in the Snorkel software package.\cite{RatnerAlexanderJandBachStephenHandEhrenbergHenryandFriesJasonandWuSenandRe2018Snorkel:Supervision}
Note that this application represents an MIL setting wherein the label extracted from a report is applied to all images in the study, but it is unknown which of these images contains the abnormality. 

The image model for this task takes a single radiograph as input and outputs the probability that this radiograph is abnormal. 
We use a 50-layer Residual Network (ResNet-50) architecture modified to emit a binary classification result on each radiograph in the training set, and leverage ImageNet-pretrained weights for model initialization.\cite{He2016,JiaDeng2009}  
The model is implemented using the Keras software package\cite{Chollet2015} and trained using the Adam optimizer with early stopping, a learning rate of $\alpha=0.0001$ with learning rate decay rate of 0.1 on plateau in the validation loss, a weight decay value of $0.005$, and the binary cross-entropy loss function.  
Images were preprocessed using histogram equalization, downsampled to 224 x 224 resolution, z-score normalized using global mean and standard deviation values computed across the dataset, and replicated over three channels (for compatibility with a model originally for RGB images) before injection into the training loop.  
Batch size was set at 60 radiographs, the maximum possible on the single 1080 Ti GPU that was used to train each model.  
Models over 30,000 training images took approximately six hours to train. 
For each exam in the test set, we compute an output score by taking the mean of the image model outputs for all radiographs in the exam, following Rajpurkar et al.\cite{Rajpurkar2017} 
Output values are reported in graphical form using ROC-AUC as a performance metric, with 95\% confidence intervals on the ROC-AUC values derived from five separate runs of each model for each dataset size using different random seeds for train-development splitting and other randomized parameters.  The held-out test set was the same in all cases.

\subsection{Intracranial Hemorrhage Detection on Computed Tomography}

The problem of rapid intracranial hemorrhage detection on computed tomography of the head (HCT) represents an important task in clinical radiology to expedite clinical triage and care.\cite{Saver2006}
To create a dataset describing a binary task of intracranial hemorrhage detection on HCT, 5,582 non-contrast HCT studies performed between the years 2001 and 2014 were acquired from our institution's PACS system.
 Each study was evaluated for a series containing 5 mm axial CT slices with greater than 29 slices and fewer than 45 slices. 
 For each study containing the requisite 5 mm axial series, the series was padded with additional homogeneous images with Hounsfield Units of 0 such that each series contained 44 CT slices.  
 The center 32 slices were then selected for automated analysis. 
 The final dataset contains 4,340 studies preprocessed in this way, and 340 of these examples are provided with a ground truth label at the scan level (cf. the slice level\cite{Lee2018}) confirmed by consensus of two radiology fellows.  
 50\% of these hand-labeled scans (170 scans) are allocated for cross validation during end model training, while the rest are used for evaluation. 
 Scan-level hand labels for the 4,000 images that support assessment of model performance using fully supervised training were provided via single-annotator reads of each report.  
 An example set of 32 CT slices for both hemorrhage and non-hemorrhage cases can be found in Fig.~\ref{fig:results-data-msk}.

The seven LFs for the hemorrhage detection task were written in Python entirely by a single radiology resident in less than eight hours of cumulative development time, using a labeled development set of 200 reports.  
Notably, these LFs programmatically combine many possible expressions for normality or hemorrhage that are drawn from this radiologist's personal experience.  
Again, probabilistic labels are generated from these LFs using tools in the Snorkel software package.\cite{RatnerAlexanderJandBachStephenHandEhrenbergHenryandFriesJasonandWuSenandRe2018Snorkel:Supervision}

As with the knee extremity application, the hemorrhage detection task fits naturally with the framework of multiple-instance labeling (MIL) because we have a single label per CT scan, each of which contains 32 image slices. 
Further, if a CT image is labeled as positive, there must exist at least one slice that contains evidence of a hemorrhage.   
However, unlike extremity radiography, in this case the different images are spatially connected along the axial dimension.  
While 3-D neural networks have recently been used for this task,\cite{Titano2018} these models often require substantially more parameters than would the attention-based approach to MIL recommended by Ilse et al.\cite{Ilse2018} 
Specifically,  Ilse  et  al.~propose an  attention-based  deep  MIL  algorithm aimed  at  improving  interpretability and increasing  flexibility by embedding  each instance (i.e. each slice) into  a  feature space and subsequently learning a weighted average off all instance (i.e. slice) embeddings corresponding to a single bag of instances (i.e. a single CT scan).  
These weights are  learned  using  a  two-layer  neural  network known as an ``attention layer." 
Classification is then performed using the attention-weighted combination of all instances  (i.e. all CT slices).

Our attention-based MIL model uses a randomly initialized ResNet-18 encoder with an output size of 50. 
Each slice is downsampled to dimensions of 224 x 224 and z-score normalized using global mean and standard deviation values computed across the dataset.  
Model training is accomplished using the stochastic gradient descent optimizer in PyTorch with a learning rate that was initially set to a high value of 0.1 and reduced upon plateaus in the validation loss, a momentum of $0.9$, and a weight decay value of $0.005$. 
Positive examples were oversampled such that the train set contained 50\% positive examples.
Batch size was set at 12 CT scans, the maximum possible using the single Tesla P100 GPU that was used to train each model.  
Training each model on the full set of images took approximately six hours.

Because only 39 of the 340 consensus-labeled examples were positive for hemorrhage, the end-model performance metrics could be sensitive to the random splitting of the test and development datasets. 
We therefore carry out a cross-validation procedure where we repeat the stratified 50-50 development set/test set split using the 340 gold labeled data points, and analyze average model performance over five trials with different random seeds, where one of the seeded operations was the development set/test set split. 

\subsection{Seizure Monitoring on Electroencephalography}

One of the most common tasks performed using EEG is epileptic seizure detection, in which an epileptologist examines large amounts of time-series data to determine whether the repeated, uncontrolled electrical discharges suggestive of seizure activity have occurred. 
Our EEG dataset comprises 30,000 pediatric EEG signals from our institution along with 9,496 EEG reports; these were randomly selected from a total of 36,644 EEG signals obtained from our institution.
Each EEG report can reference multiple signals, and each signal can be referenced by multiple EEG reports.
Each signal is annotated by an EEG technician with onset times of possible seizures, which we treat as full hand-labeled supervision for a machine learning model for seizure detection.
In order to ensure consistency across exams, each of which could have a unique sensor alignment, we use only signals from the 19 electrodes in the standard 10-20 International EEG configuration, which form a subset of the electrodes deployed to every patient at our institution.  
Voltage readings from each channel are sampled at 200 Hz.  
For the sake of simplicity, we transform the seizure onset detection problem into a clip-level classification problem over 12-second clips of the full time series.
Our model maps an input $x \in R^{2400x19}$ to a single output indicating the probability of seizure onset in that clip.  
An example set of EEG signals can be found in Fig.~\ref{fig:results-data-eeg}.

The eleven LFs used to determine if a given EEG clip contains a seizure onset were written collaboratively by a clinical neurologist and a postdoctoral computer scientist over the course of less than eight cumulative hours of dedicated time, using a labeled development set of 200 reports.  
These LFs simultaneously leverage the structure of the EEG report along with unstructured information contained in the raw text, which can often cover several paragraphs.  
The Snorkel software package is then used to create probabilistic labels for each report.\cite{RatnerAlexanderJandBachStephenHandEhrenbergHenryandFriesJasonandWuSenandRe2018Snorkel:Supervision}
Due to the length of these reports and their highly variable structure, the LFs for this application represent a particularly compelling example of how domain-specific knowledge can be used to inform heuristic development.  

  Because these reports refer to entire signals rather than to specific clips, a small hand-labeled EEG clip dataset for cross-validation and end-model evaluation was created by a pediatric clinical neurologist with 10 years of experience.  
  This dataset contains 350 12-second clips representative of seizure onset; 100 of these positive examples are allocated to a development set for cross-validating the end model and 250 are allocated to the test set for final evaluation.  Sets used for both cross-validation and evaluation were made to contain an 80\% fraction of clips without seizure onset by randomly sampling clips from signals confirmed to contain no evidence of seizure.  

We use a densely connected inception architecture inspired by Roy et al.\cite{Roy2018} for seizure onset detection.
This modeling approach combines the most compelling aspects of the InceptionNet\cite{szegedy2015going} and DenseNet\cite{Huang2016} architectures, namely the extraction of convolutional features at multiple granularities at each layer combined with concatenation of each filter bank with all of those preceding it.  
In order to address the issue of extreme class imbalance caused by the low frequency of clips containing seizure onset even in EEG signals where a seizure is known to be present, we use a simple filtering process analogous to common techniques used in the information extraction literature for candidate extraction for weakly supervised models.\cite{RatnerAlexanderJandBachStephenHandEhrenbergHenryandFriesJasonandWuSenandRe2018Snorkel:Supervision}
Specifically, we construct a candidate extractor that is a simple three-layer neural network operating on a set of 551 features reported to be useful for seizure detection from the literature.\cite{Boubchir2017}  
This candidate extractor, which is trained on the development set, is executed over all signals with an associated report that is weakly labeled as containing a seizure to provide clips that are positive for seizure onset with high probability, while negative clips are randomly sampled from signals with no associated positive report.
The candidate extractor cutoff was tuned using the development set such that precision was 100\% and recall was 15\%; because we have such a large unlabeled dataset, we are able to optimize for this high level of precision while still obtaining a large set of positive examples.

  Model training is accomplished using the Adam optimizer in PyTorch, learning rate was initially set to a value of 1e-6 and reduced upon plateaus in the validation loss, and a dropout probability of $0.2$ was applied to the last layer. 
  Plentiful negative examples were undersampled such that the train set contained 50\% positive examples.
  Batch size was set at 10 EEG signals, the maximum possible using the single Tesla P100 GPU that was used to train each model.  Training each model on the full set of signals took approximately 12 hours 
 
  Performance values are reported in graphical form using ROC-AUC as a performance metric, with 95\% confidence intervals on the ROC-AUC values derived from five separate runs of each model for each dataset size using different random seeds for model initialization, train-development splitting, and other randomized parameters.  The held-out test set was the same in all cases.  
  
  \subsection{Statistical Analyses}
  Relevant statistical tests performed in this work are two-tailed DeLong non-parametric tests to evaluate the equivalence of ROC-AUC values.  All ROC-AUC values are computed using the entire test sets described above, which indicates sample size for each.  These tests were implemented using the R package pROC accessed via the Python package rpy2.

%% file: sections-supp/statements.tex


\begin{addendum}
\item [Data Availability] Protected Health Information restrictions apply to the availability of the clinical datasets presented here, which were used under Institutional Review Board approval for use only in the current study, and so are not publicly available.  We have, however, made available the labeling functions used for each application as well as a full demo of the cross-modal weak supervision pipeline using a small public dataset containing chest radiographs and text reports in an online repository.\footnote{\demourl}  Raw data for Figures 3, 4, and 5 is available from the authors.

\item [Code Availability Statement] All code used for this study uses open-source Python packages.  We have made the labeling functions themselves as well as a full demo of the cross-modal weak supervision pipeline using a small public dataset containing chest radiographs and text reports in an online repository (see Footnote).  Technical background and usage information on the Snorkel software package used heavily in this work can be found at snorkel.stanford.edu.

 \item We gratefully acknowledge the support of DARPA under Nos. FA87501720095 (D3M) and FA86501827865 (SDH), NIH under Nos. N000141712266 (Mobilize), 1U01CA190214, and 1U01CA187947, NSF under Nos. CCF1763315 (Beyond Sparsity) and CCF1563078 (Volume to Velocity), ONR under No. N000141712266 (Unifying Weak Supervision),the National Library of Medicine under No. R01LM01296601,  the Stanford Child Health Research Institute under No. UL1-TR001085, the Stanford Human-Centered Artificial Intelligence Seed Grant, the Moore Foundation, NXP, Xilinx, LETI-CEA, Intel, Google, NEC, Toshiba, TSMC, ARM, Hitachi, BASF, Accenture, Ericsson, Qualcomm, Analog Devices, the Okawa Foundation, American Family Insurance, and members of the Stanford DAWN project: Intel, Microsoft, Teradata, Facebook, Google, Ant Financial, NEC, SAP, and VMWare.   J.D. was supported by the Intelligence Community Postdoctoral Fellowship, A.R. was supported by the Stanford Interdisciplinary Graduate Fellowship, and H.S. was supported by an RSNA Research Fellow Grant from the Radiological Society of North America and the NIH No. T32-CA009695 (Stanford Cancer Imaging Training Program). The U.S. Government is authorized to reproduce and distribute reprints for Governmental purposes notwithstanding any copyright notation thereon. Any opinions, findings, and conclusions or recommendations expressed in this material are those of the authors and do not necessarily reflect the views, policies, or endorsements, either expressed or implied, of DARPA, NIH, ONR, or the U.S. Government.

\item[Competing Interests] The authors declare no competing interests.
 
\item [Author Contributions]  A.R., J.D., D.R., and C.R. conceptualized the study.  J.D., A.R., R.G., N.K., K.S., D.R., and C.R. contributed to experimental design while J.D., A.R., N.K., and K.S. wrote computer code and performed experiments.  M.L., R.G., C.L.M., M.M., H.S., and D.R. provided labeled data and clinical expertise for the study.  R.G., N.K., C.L.M., and J.D. contributed to labeling function development and tuning.   J.D. and A.R. prepared the manuscript.  All authors contributed to manuscript review.  (*) and (**) indicate equal contributions.    
 \item[Material and Correspondence] Correspondence and requests for materials
should be addressed to J.D. ~(email: jdunnmon@stanford.edu). 
\end{addendum}

%% file: sections-supp/supp-mat.tex

\section*{Extended Data}



\begin{table}[!tbh!]
\centering
\begin{tabular}{|c|c|c|c|c|}
\hline
\hline
Dataset  & CXR & EXR & HCT & EEG  \\
\hline
Large FS vs. DP & 0.3821 & 0.0225 & 0.6223 & 0.0000 \\
\hline
Medium FS vs. DP  & 0.0004 & 0.3178 & 0.0006 & 0.2685 \\
\hline
\hline
\end{tabular}
\label{table:p-values}
\caption{P-values from the two-tailed DeLong non-parametric test comparing ROC curves from median models.  These results demonstrate that median CXR and HCT models supervised with cross-modal data programming (DP) are not statistically distinguishable from those supervised with the Large fully supervised (i.e. hand-labeled) set (p$>$0.3), and are significantly different (p$<$0.001) than those supervised with the Medium fully supervised set (see Table \ref{table:app-table} for dataset size definitions).  Similarly, for EEG and EXR, median models supervised with cross-modal data programming were statistically different than those supervised using the Large supervised set (p$<$0.05), but not distinguishable from those trained using the Medium fully supervised set (p$>$0.25).}
\end{table}